\documentclass[conference]{IEEEtran}
\IEEEoverridecommandlockouts
\usepackage{cite}
\usepackage{amsmath,amssymb,amsfonts}
\usepackage{amsthm}
\usepackage{url}
\usepackage{graphicx}
\usepackage{textcomp}
\usepackage{xcolor}
\usepackage{comment}
 \usepackage{booktabs}
\usepackage[numbers]{natbib}
\usepackage{algorithm, algpseudocode}
\def\BibTeX{{\rm B\kern-.05em{\sc i\kern-.025em b}\kern-.08em
    T\kern-.1667em\lower.7ex\hbox{E}\kern-.125emX}}
\begin{document}

\title{FAST-ME: Foundation-aware Adaptive Stopping for  Motion Estimation for Efficient IoT Video Analysis}

\author{
\IEEEauthorblockN{1\textsuperscript{st} Kakia Panagidi}
\IEEEauthorblockA{\footnotesize
\textit{Department of Informatics \& Telecommunications} \\
\textit{National and Kapodistrian University of Athens} \\
Athens, Greece \\
kakiap@di.uoa.gr}
\and
\IEEEauthorblockN{2\textsuperscript{nd} Stathes Hadjieftymiadis}
\IEEEauthorblockA{\footnotesize
\textit{Department of Informatics \& Telecommunications} \\
\textit{National and Kapodistrian University of Athens} \\
Athens, Greece \\
shadj@di.uoa.gr}
\thanks{\textit{*Both authors contributed equally to this research.}}
}

\maketitle

\begin{abstract}

In modern multimedia systems, efficient video processing is critical, especially in resource-constrained environments such as IoT-based camera networks, autonomous platforms, and wireless sensor multimedia systems. A key bottleneck in video compression and understanding is block motion estimation (ME) — a process that remains computationally expensive despite the development of fast search techniques. This research work introduces an Optimal Stopping Theory algorithm applied on the block motion estimation based on the assessment of spatiotemporal differences within and among frames of the video sequence and proposes a novel semantic-aware motion estimation framework that integrates Foundation Models (FMs) with the OST-based decision process. By leveraging pretrained visual transformers such as ViT and SAM, the framework extracts semantic attention scores that indicate the importance of motion within specific spatial regions. These scores are fused with traditional distortion-based metrics (e.g., SAD) to guide a hybrid stopping criterion that jointly considers motion magnitude and semantic relevance. The result is an adaptive algorithm that stops early in redundant regions but continues searching where motion is semantically significant.Experiments have been conducted to compare the proposed solution with the well-used approached in the literature on benchmark and multimodal video datasets. The proposed method achieves up to a 99\% reduction in computation, with minimal accuracy loss and improved semantic coverage. The evaluation results highlight the benefits of bridging low-level motion analysis with high-level semantic reasoning, offering a promising direction for efficient multimodal video understanding in next-generation smart systems.

\end{abstract}





\maketitle

\section{Introduction}
Wireless Sensor Multimedia Networks (WSMNs) \cite{wsmn15} nowadays attract significant attention because of the variety of applications in which they can be applied such as traffic congestion, environmental, user monitoring and recording unusual events. The growth estimation of WSMNs between 2023 and 2028 is close to USD $110.73$ billion driven by the demand of the use of wireless multimedia in car technology, vehicle automation, Internet of Things (IoT) connectivity and infrastructures. One of the features, which is energy consuming in WSMNs, is multimedia streaming. Multimedia streaming is the process of sending and delivering multimedia content to end users or to the fixed infrastructure, where it will pass through further processing. Motion Estimation (ME) has played an important role in video processing. It is usually applied to block matching algorithms to choose the best motion vector. The two neighboring frames are searched to find the displacement of the same object in the video image. Motion estimation (ME) techniques aim at deducing displacement vectors for objects or image attributes between two consecutive frames. The object motion encoder calculates the motion between the current and the reference frame.

There are two distinct phases of block matching method: block partitioning and block searching. The block partitioning scheme is concerned with dividing the original image frame into non-overlapping regions and it performs by using the fixed size or variable size methods.  Block matching builds on this concept by dividing the current frame into a grid of “macro blocks" \cite{Tekalp95}. Each macro block is then compared to a corresponding block and its neighboring blocks in the previous frame to determine a movement vector. This vector indicates how the macro block has shifted from one location to another in the earlier frame. By calculating these movements for all the macro blocks in a frame, the overall motion in the current frame can be estimated. The search area for identifying the best matching macro block is limited to a range of "p" pixels in all directions around the corresponding block in the previous frame. This ‘p’ is called as the search parameter. Larger motions require a larger p, and the larger the search parameter the more computationally expensive the process of motion estimation becomes . Usually the macro block is taken as a square of side 16 pixels, and the search parameter p of pixels \cite{sayood2006}. The block search mechanism is the process of locating the block in the destination frame that best matches the block in the frame using a specific matching criterion. 

Different distortion measures are used to find the best match for a desired macro block in the entire motion estimation process like the mean absolute error (MAE), mean squared error (MSE) and the sum of absolute differences (SAD). Equation \ref{eq:sad} provides the Sum of Absolute Differences (SAD) for two frames, i.e current and reference frame,  of size  $N \times N$  as:
\begin{equation}
\text{SAD} = \sum_{i=1}^{N} \sum_{j=1}^{N} | B_c(i, j) - B_r(i, j) |
\label{eq:sad}
\end{equation}

Where  $B_c(i, j)$  and  $B_r(i, j)$  are the pixel intensities at position $(i, j)$ in the current frame and the reference frame, respectively and  $N$  is the size of the block (e.g., $16 \times 16$).  An Example (for a 2x2 block) is shown below: 
\begin{multline}
\text{SAD} = |B_c(1,1) - B_r(1,1)| + |B_c(1,2) - B_r(1,2)| 
+ |B_c(2,1) - B_r(2,1)| + |B_c(2,2) - B_r(2,2)|
\end{multline}

ME operates on the premise that patterns representing objects and backgrounds in a video frame shift within the frame to align with their counterparts in the subsequent frame. For example a Full Search approach \cite{BarjatyaMPEG04} is the most computationally expensive block matching algorithm of all because it calculates the cost fundamentalction at each possible location in the search window, while Diamond Search (DS) \cite{BarjatyaMPEG04} follows a search point pattern of a diamond and it is more efficient in computation resources. Based on the ME processing, we introduce an adaptive approach that monitors the spatiotemporal differences of the motion vectors computed and stops early if no differences occurred. The idea is simple: the model overviews the observations of the current and the predicted frames and stops when the best possible match of differences occurred avoiding unnecessary evaluations. In scenarios where inter-frame changes are minimal, continuing motion estimation may be computationally redundant, such as in a static parking lot video. The proposed method applies the principles of Optimal Stopping Theory, particularly inspired by the ‘burglar problem’ formulation \cite{CHRISTENSEN2020}. 

However, traditional OST-based strategies are fundamentally content-agnostic, i.e. they focus only on pixel-level differences and do not consider semantic relevance of motion. For instance, in surveillance footage, a small motion in a person’s face may be far more important than larger movements in the background. This leads to a critical limitation: uniform motion treatment regardless of content importance. To overcome this limitation, in this paper we further propose a hybrid framework, i.e. FAST-ME: Foundation-aware Adaptive Stopping for Targeted Motion Estimation, that integrates Foundation Models (FMs) — such as the Vision Transformer (ViT), Segment Anything Model (SAM), and CLIP — with the OST-based motion estimation paradigm. These large-scale pretrained models offer powerful capabilities to extract semantic attention, object saliency, and multimodal understanding from video frames. Leveraging their output, the hybrid approach dynamically weighs the motion estimation cost using both distortion and semantic saliency, enabling the system to prioritize motion in semantically meaningful regions while avoiding unnecessary processing in unimportant areas.

The ultimate goal of this hybrid approach is to: i) incorporate content-awareness into the block matching decision-making process; ii) enhance the adaptivity of the stopping rule using semantic guidance; iii) bridge the gap between statistical optimality (OST) and semantic perception (FMs) and iv) enable scalable multimodal video analysis in bandwidth- or power-limited systems. This semantic-aware OST framework is tested on standard and multimodal video datasets, showing improvements in computation efficiency, semantic preservation, and compression quality. This positions FAST-ME at the intersection of low-level motion processing and high-level video understanding, a key theme in emerging multimedia systems powered by foundation models. The model is compared with the well-known approaches in block estimation as described in \cite{BarjatyaMPEG04}. The results demonstrate improved efficiency, enhanced semantic preservation, and strong PSNR performance.

The rest of the paper is organized as follows: Section~\ref{sec:relatedwork} reviews the literature on motion estimation and foundation models. Section~\ref{sec:adaptive_me} presents the original OST-based adaptive motion estimation approach. Section~\ref{sec:hybrid} introduces the FAST-ME framework. Experimental results are presented in Section~\ref{sec:experiments}, and Section~\ref{sec:conclusion} concludes the paper with future directions.

\begin{table}[h]
    \caption{Nomenclature}
    \label{tab:nomenclature}
    \centering
    \begin{tabular}{p{2cm} p{5cm}}
        \toprule
        \textbf{Symbol} & \textbf{Definition} \\
        \midrule
        $\alpha$ &  blending factor\\
        $b$ & Block size (typically 16×16 or 8×8  in practice).  \\
        $B_c(i, j)$ & Intensity of the pixel at position \((i, j)\)  in the current block.  \\
        $B_r(i, j)$ & Intensity of the pixel at position \((i, j)\) in the reference (candidate) block.  \\
        $\delta$ & A predefined threshold representing the acceptable  probability of finding  a better match (threshold base).  \\
        $\mathbb{E}[V_{n-1}]$ & Expected value of continuing the search.  \\
        $\theta$ & Rate parameter.  \\
        $F_Y(Y_k)$ & Cumulative Distribution Function (CDF)  of SAD values at time \(k\).  \\
        $p$ & Number of pixels defining the search area  on all four sides.  \\
        $SAD$ & Sum of Absolute Differences (SAD), used as  a cost function for block matching.  \\
        $T$ & Parameter controlling the strictness  of the stopping rule.  \\
        $\tau^*$ & Optimal stopping time (the index  at which the search stops).  \\
        $V_n(y)$ & Value function at step \(n\),  given an observation \(y\).  \\
        $W(y)$ & Immediate reward if the search  stops at \(y\).  \\
        $Y_k$ & SAD value at time \(k\).  \\
        $Y_{\infty}$ & Limiting value of the motion vectors.  \\
        \bottomrule
    \end{tabular}
\end{table}

\section{Related Work and Contribution} \label{sec:relatedwork}
The basic notion in motion estimation is to identify the block of objects existing in consecutive frames crating a group of objects. The idea behind the block matching is to divide a frame in macroblocks which are compared with the previous frame to create a vector demonstrating the movement of the objects between blocks. 
\subsection{Related Work}
Starting  defining the ground truth of motion estimation approaches, authors in  \cite{BarjatyaMPEG04} present the commonly used block matching algorithms used for motion estimation. All seven approaches are studied and performance evaluated, i.e. Full/Exhaustive Search, Three step Search (TSS), New Three Step Search (NewTSS), Simple and Efficient TSS, Four step search (FSS), Diamond search(DS) and Adaptive Rood pattern Search. Going a step further authors in \cite{lai2019}, \cite{Soroushmehr22}, \cite{Purnachand2012}, \cite{Cai09},\cite{Kuo06} and \cite{Tsai06}  are proposing a fast estimation approach of the original approaches based on updates on the decrease of the $p$ neighbors' pixels, rotation of shape in x-axis and introduction of new indicators like the correlation between the referenced blocks. For example in \cite{Cai09} a simplified reference frame decision strategy named as spatial neighbor searching scheme is introduced to speed up the
selection of the optimal reference frame, while in \cite{Tsai06} and \cite{Kuo06} proposed alternatives in motion field distribution or a three-dimensional predict hexagon search method to reduce the computational load.

A threshold approach is introduced in  \cite{Purnachand2012} where the threshold can be chosen adaptively by averaging the costs of all the previous inter prediction coding units in the first frame (except intra frame) of Group of Pictures in HEVC encoder. A threshold for dynamic video encoding applied in a Groups-of-Pictures  based on Optimal Stopping Theory are presented in \cite{panagidi18} identifying scene changes. 
Other works \cite{True2021} \cite{He2016}, \cite{Dai2016} and \cite{Zhu2017} in Motion Estimations are mostly in the domain of object detection, where object detection techniques are used together with motion estimation action in paraller. Several use of neural networks structures are proposed in order to identify efficiently a referenced object, e.g. CNN networks \cite{True2021}, ResNet 50 and 101 \cite{He2016} and RFCN head networkrs \cite{Dai2016} and \cite{Zhu2017}. The training of a neural network is going back to the limitations due to the size of datasets training, the need for resources for the training and the problem of the hyper-tuning. 

\subsection{Foundation Models and Semantic-Aware Video Analysis}

Recent advances in \textit{Foundation Models (FMs)} have significantly reshaped the landscape of video understanding and multimodal analysis. These large-scale pretrained models, such as \textbf{CLIP}~\cite{radford2021clip}, \textbf{SAM}~\cite{kirillov2023sam}, and \textbf{ViT}~\cite{dosovitskiy2021vit}, enable generalization across tasks like object detection, segmentation, action recognition, and language grounding with minimal supervision. In the context of video analysis, \textbf{Vision Transformers (ViT)} and their derivatives (e.g., TimeSformer~\cite{bertasius2021timesformer}, VideoMAE~\cite{tong2022videomae}) have demonstrated impressive performance in capturing spatiotemporal dynamics. Meanwhile, \textbf{CLIP} enables image-text alignment, supporting zero-shot classification, retrieval, and semantic search in video streams~\cite{lei2021clipbert, luo2023semsearch}. \textbf{SAM}, on the other hand, provides dense and prompt-free instance segmentation that can be integrated into video segmentation, saliency detection, and scene analysis tasks~\cite{kirillov2023sam}.

Despite these advancements, most traditional video coding pipelines still rely on pixel-level heuristics such as Sum of Absolute Differences (SAD) and Mean Squared Error (MSE) for motion estimation. These methods fail to consider semantic content or the relevance of motion within the scene. Emerging efforts such as CLIP-Compress~\cite{li2023clipcompress} and content-aware bit allocation~\cite{chen2023contentaware} begin to bridge this gap by using foundation model outputs to guide perceptual quality during compression. However, these works typically operate at the frame or scene level rather than integrating semantic priors into the \textit{motion estimation loop}. Our work introduces a hybrid framework that integrates semantic attention from foundation models directly into block-level motion estimation through Optimal Stopping Theory (OST). This enables a principled, adaptive mechanism to reduce unnecessary computation while preserving semantically important motion, paving the way for intelligent and resource-efficient video coding.

\subsection{Contribution}
This work proposes a hybrid motion estimation framework that integrates semantic guidance from foundation models into the optimal stopping theory-based block matching process. An adaptive motion estimation (ME) approach is introduced based on changes in motion vector calculations using Optimal Stopping theory rule in order to ensure the efficient use of the resources and the quick adaptations to changes:
	\begin{enumerate}
        \item \textit{an OST rule for time-optimized motion estimation named as Adaptive Motion Estimation (ME) }, 
    \item \textit{A hybrid cost function that fuses traditional distortion measures (e.g., SAD) with semantic attention extracted from FMs, enabling content-aware motion estimation.}
    \item \textit{A reformulated OST stopping rule that adapts dynamically to both motion intensity and semantic saliency, prioritizing meaningful motion over background noise.}
    \item \textit{A comparative evaluation across multiple video datasets, demonstrating that the proposed method achieves competitive motion vector accuracy while reducing unnecessary computation. }
	\end{enumerate}

This integration advances the field toward semantically-aware, computation-efficient, and scalable video processing architectures, making it highly relevant to emerging applications in multimodal video understanding, edge intelligence, and IoT vision systems.

\section{Adaptive Motion Estimation Model Architecture}\label{sec:adaptive_me}
\subsection{Preliminaries}

In ME problem, we’re trying to determine the best motion vector by comparing blocks in consecutive frames. We can recall three main categories for ME problem in the literature: 
\begin{itemize}
    \item A Full Search (FS) algorithm divides the current frame into 16x16 blocks. For each block, FS searches within the range $p$ pixels in the reference frame computing SAD for each candidate position, and selects the one with the minimum SAD;
    \item The Diamond Search algorithm using a diamond-shaped pattern to search for the best match in the reference frame. DS typically starts with a large search range and progressively reduces the search area progressively by checking fewer candidate positions, based on previous results;
    \item Three-Step Search (TSS) works by iteratively reducing the search area in three stages: i) it starts by searching a wide range; ii) it narrows the search to a smaller range based on the results from the first stage; and iii) it performs a finer search around the best candidate found.
\end{itemize}

\begin{figure}
    \centering
    \includegraphics[width=0.75\linewidth]{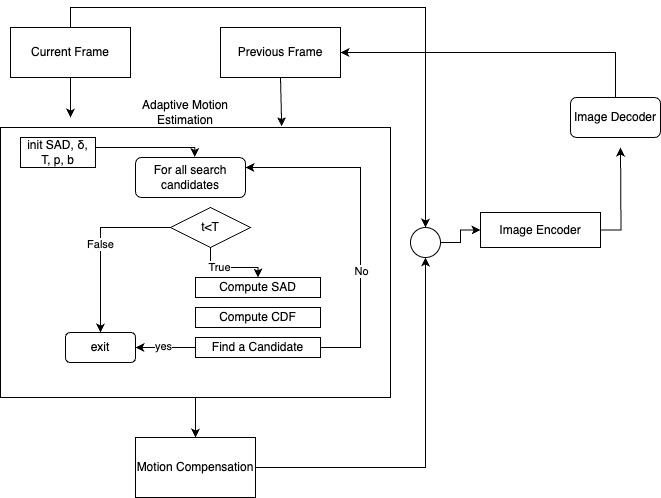}
    \caption{Adaptive ME Model diagram}
    \label{fig:AME}
\end{figure}
Let consider an Adaptive ME model as shown in Figure \ref{fig:AME} that overviews the SAD values during the motion estimation process. The goal is to identify the best matching block between the $B_c(i, j)$ and the $B_r(i, j)$ frame by searching for the minimum SAD in order to determine whether continuing to search for a better match is likely to result in a better match or not. In other words, instead of exhaustively searching through all candidate blocks (as in FS), an optimal stopping rule is proposed to stop early once a “good enough” match is found.

\begin{figure}
    \centering
    \includegraphics[width=0.7\linewidth]{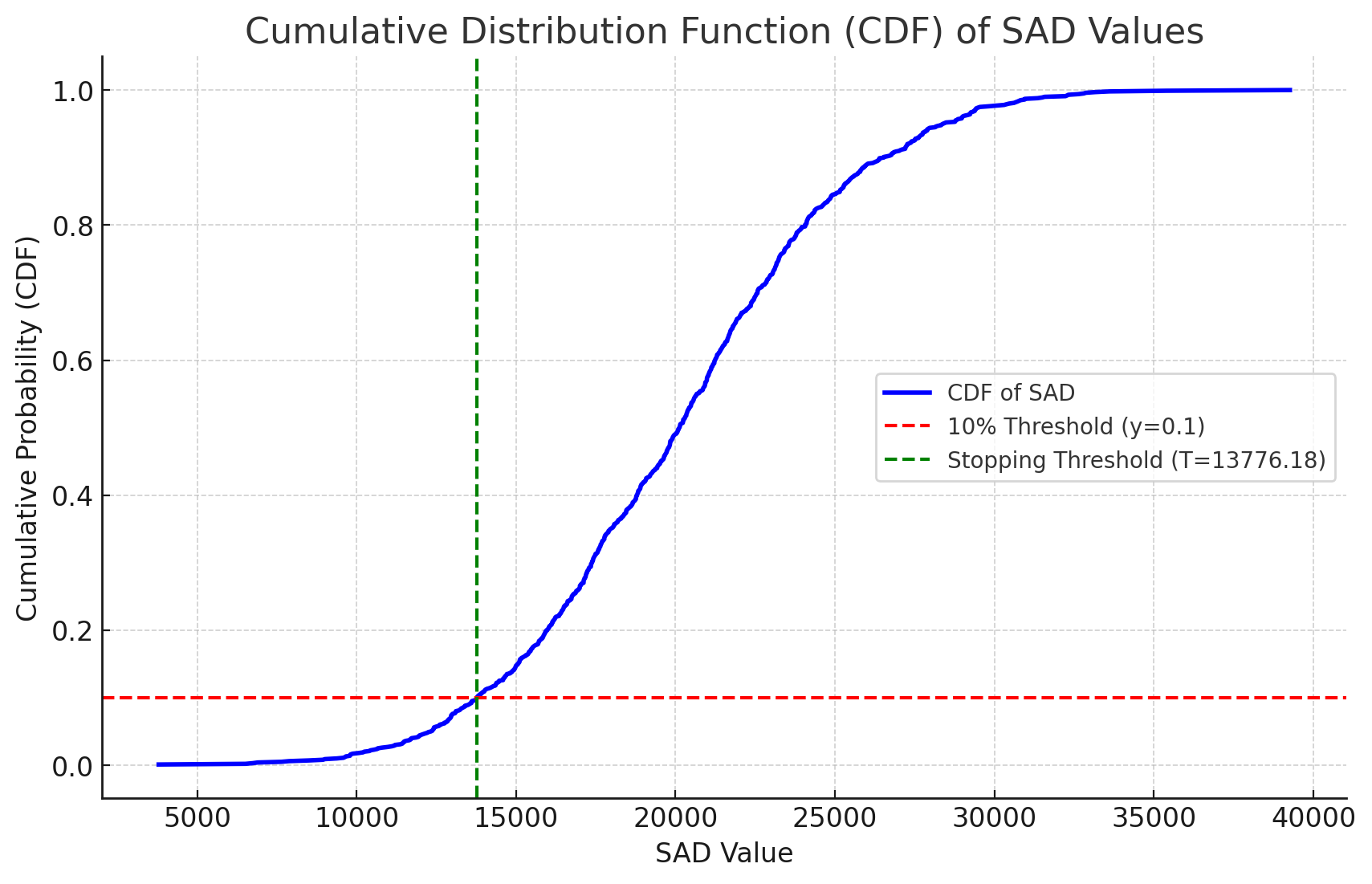}
    \caption{Cumulative Distribution $y(t)$ of SAD in different $T$}
    \label{fig:CDF1}
\end{figure}

\begin{figure}
    \centering
    \includegraphics[width=0.7\linewidth]{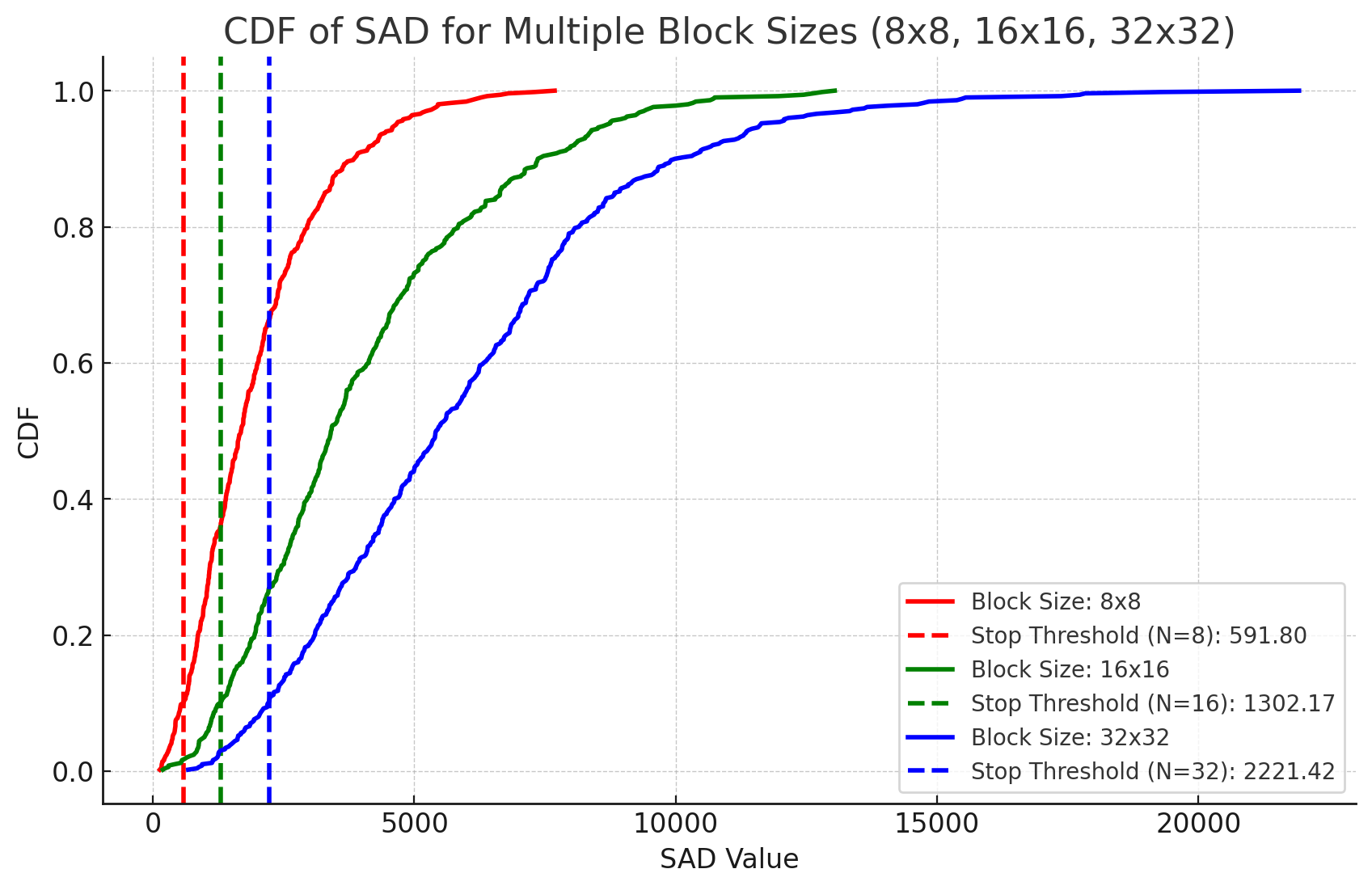}
    \caption{Cumulative Distribution $y(t)$ of SAD in different Block Size}
    \label{fig:CDF_all}
\end{figure}

\subsection{OST hypothesis}
Let us assume that  $Y_{1}, Y_{2}, \dots, Y_{n}$  be the observed SAD values, representing the cost of matching blocks at each step  $t = 1, 2, \dots, n$ . The CDF  $F_Y(y)$  is given by:
\begin{equation}
   F_Y(y) = P(Y \leq y) = \frac{1}{N} \sum_{i=1}^{N} \mathbb{I}(Y_i \leq y)
\end{equation}

$F_Y(y)$ is presented in Figure \ref{fig:CDF1}, where the blue curve represents the cumulative distribution function (CDF) values of the SAD and the red dashed line corresponds to threshold $T$, in this case namely as  $10\%$ of CDF, and the green line indicates the optimal threshold $T$, in which $10\%$ of the SAD values lie below this point. This threshold can be used as a decision point to stop searching for better motion vectors when a SAD value below $T$ is found. The impact of different block sizes $N$ on $F_Y(y)$ is shown in Figure \ref{fig:CDF_all}. The dashed vertical lines represent the optimal stopping thresholds $T=10\%$ for each block size  $N \in [8, 16, 32]$.  $F_Y(y)$ is larger for larger blocks as expected with a shift to the right. 

The aim of the model is to stop when the likelihood of finding a lower SAD value decreases below a certain threshold  $\delta$ .This threshold is defined as: $F_Y(Y_k) \geq 1 - \delta$ where  $Y_k$  is the current SAD value at step  $k$  and  $\delta$  is a small probability value (e.g.,  $\delta = 0.05$  or  $\delta = 0.01$ ). If the $F_Y(y)$ at the current SAD value is large enough (i.e., the probability of finding a better match is low), we stop and accept the current motion vector. The optimal stopping rule would then be the point where: $F_Y(Y_k) \geq 1 - \delta$ which means that the probability of finding a better match is smaller than  $\delta$ , and therefore it is no longer beneficial to continue the search. The stopping rule based on the $F_Y(y)$ can be written as follows:
\begin{equation}
    \tau^* = \min\left\{ k \mid F_Y(Y_k) \geq 1 - \delta \right\}
\end{equation}
where 
\begin{align*} 
F_Y(Y_k) = 1 - e^{-\theta Y_k} \geq 1 - \delta
\end{align*} 
Solving for $Y_k$, we get:
\begin{equation}
 e^{-\theta Y_k} \leq \delta \implies Y_k \geq -\frac{\log(\delta)}{\theta}
\end{equation}

\subsection{Optimal Stopping rule and $\delta$}

The rule is optimal in the sense that it minimizes the expected number of evaluations while ensuring that the SAD is as small as possible. By using the CDF, we incorporate statistical information about the distribution of SAD values, making the stopping decision more informed and efficient.

Consider the following inequality:
\begin{equation}
    \limsup_{n \to \infty} F_{Y_n}(Y_n) \leq Y_{\infty}
\end{equation}
where \(Y_{\infty}\) is the limiting value of the motion vectors.

We aim to identify \(\delta\) so that the optimal stopping time \(\tau^*\) is given by:
\[
\tau^* = \min \left\{ k \mid F_{Y_k}(Y_k) \geq 1 - \delta \right\}
\]
Since the stopping condition requires \(F_{Y_k}(Y_k) \geq 1 - \delta\), we relate this to the \(\limsup\) of the CDFs:
\[
1 - \delta \leq \limsup_{n \to \infty} F_{Y_n}(Y_n) \leq Y_{\infty}
\]
Rearranging the inequality, we get:
\[
1 - \delta \leq Y_{\infty}
\]
Solving for \(\delta\), we have:
\[
\delta \geq 1 - Y_{\infty}
\]
To ensure that the stopping criterion satisfies \(\limsup_{n \to \infty} F_{Y_n}(Y_n) \leq Y_{\infty}\), the threshold \(\delta\) must satisfy:
\[
\delta \geq 1 - Y_{\infty}
\]

\begin{figure}
    \centering
    \includegraphics[width=0.6\linewidth]{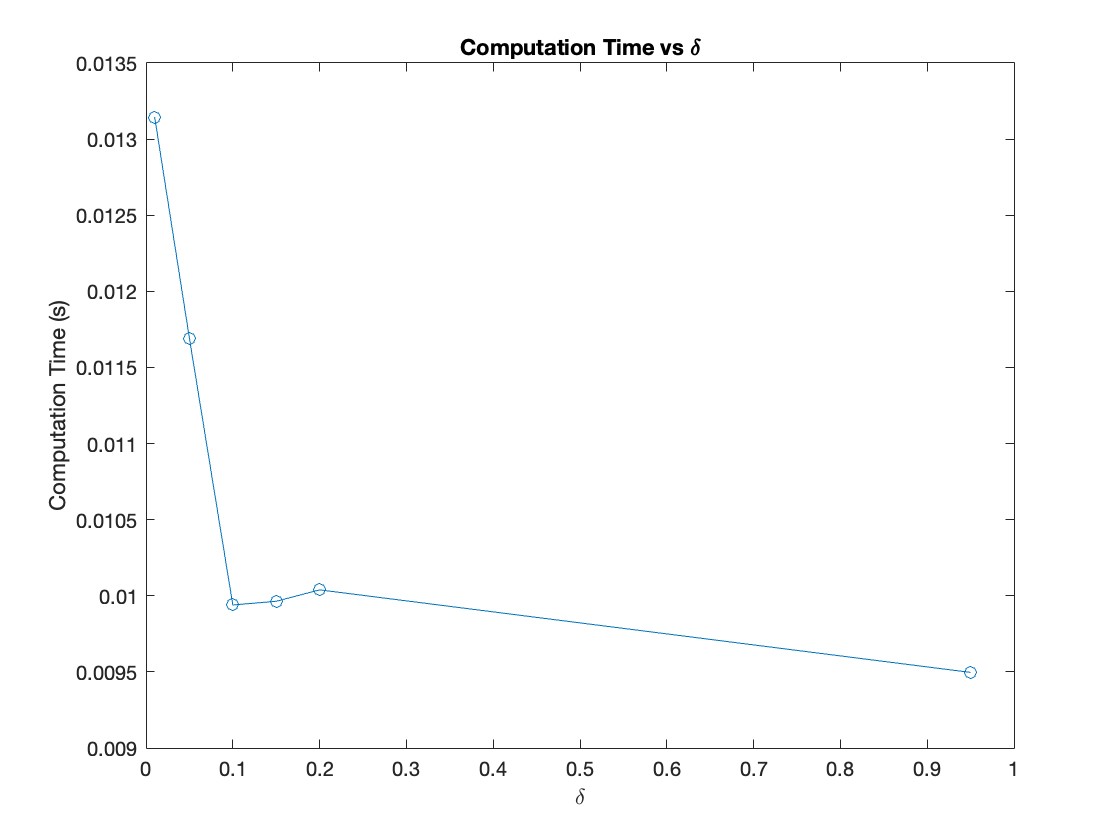}
    \caption{Computation time vs $\delta$}
    \label{fig:computation_delta}
\end{figure}

The threshold  $\delta$  can be tuned based on the specific application and the trade-off between computation time and the accuracy of the motion vector. For example, if accuracy is more important, we might set a lower  $\delta$ , allowing for a longer search. Conversely, if speed is a priority, a higher  $\delta$  might be used, stopping earlier. Giving an example, let assume  $Y_{\infty} = 0.95$, then $\delta \geq 1 - 0.95 = 0.05$. This implies that we should set the stopping condition $F_{Y_k}(Y_k) \geq 0.95$ to ensure that the stopping time $\tau^*$ is optimal according to the principles of optimal stopping theory. Figure \ref{fig:computation_delta} shows how long it takes for a motion vector model based on OST to finish for each different values of $\delta \in [0.01, 0.05, 0.1, 0.15, 0.9]$ . Higher $\delta$ values reduce computation time.

\subsection{Determining the optimal Stopping rule}

The existence of this stopping rule is guaranteed under the assumption that the SAD values follow a well-defined probability distribution. SAD values are non-deterministic and can be modeled as random variables i) due to unknown image content (the content of the image blocks are unknown before hand); ii) noise and uncertainty (image blocks are often affected by noise like sensor noise, compression techniques, etc.). The CDF is a non-decreasing function, so there will always be a point where $F_Y(Y_k) \geq 1 - \delta$ , ensuring that we can always find a stopping time. 

Let assume that SAD follows an Exponential distribution with density
\begin{equation}
    f(x \mid \theta) = \theta e^{-\theta x} \cdot \mathbb{I}(x \geq 0)
\end{equation}
where $\theta > 0$ is the rate parameter, and $\mathbb{I}(x \geq 0)$ is the indicator function that ensures x $\geq$ 0.
We consider the SAD as an exponential distribution as long as "time" in our model can be seen as the number of SAD measurements; when a big SAD is computed we stop, just like the time till someone notices an earthquake. We can observe that while we are searching through motion vectors, most SADs are large (bad matches), and only a few are small (good matches). The CDF of SAD values might have a right-skewed shape. The exponential distribution fits this behavior since it places more weight on small values and quickly “decays” for larger values like 
\begin{equation}
    f_Y(y) = \theta e^{-\theta y}, \quad y \geq 0
\end{equation}

If we assume that the SADs $Y_1, Y_2, \ldots, Y_N$ are i.i.d. samples from an $\text{Exponential}(\theta)$ distribution, the stopping rule becomes:
\begin{equation}
\tau^* = \frac{1 - e^{-\theta \tau^*}}{\theta}
\end{equation}


\begin{proof}

Let  $V_n(y)$ be the expected value of the future payoff starting from  $Y_1 = y $ at time  $n $.
The Bellman equation for the value function  $V_n(y) $ is:
\begin{equation}
    V_n(y) = \max \left\{ W(y), \mathbb{E}[V_{n-1}(Y_1)] \right\} 
\end{equation}
 
where:$W(y) $ is the immediate reward if we stop at y and $\mathbb{E}[V_{n-1}(Y_1)] $ is the expected value of continuing to search.

The distribution of the next SAD  $Y_1 $ is exponential with parameter  $\theta $. So, the expected reward  $\mathbb{E}[V_{n-1}(Y_1)] $ is:
\begin{equation}
\mathbb{E}[V_{n-1}(Y_1)] = \int_0^\infty V_{n-1}(y) \, \theta e^{-\theta y} \, dy 
\end{equation}
Since  $V_{n-1}(y) $ is recursively defined as:
\begin{equation}
 V_{n-1}(y) = \max \left\{ W(y), \mathbb{E}[V_{n-2}(Y_1)] \right\} 
\end{equation}

To explicitly compute the integral and derive the threshold for y at each stage, Bellman equation for the value function is applied. The goal is to solve the equation for the threshold $\tau^*$ where it is optimal to stop and accept the current SAD value rather than continue the search.

Let $V_n(y)$ be the value function at step $n$ starting from an observation $y$. The Bellman equation is:
\begin{equation}
V_n(y) = \max \left\{ W(y), \mathbb{E}[V_{n-1}(Y)] \right\}
\end{equation}
where:$W(y) = y$ is the immediate reward if we stop at state $y$ and $\mathbb{E}[V_{n-1}(Y)]$ is the expected reward if we continue searching.

The expectation $\mathbb{E}[V_{n-1}(Y)]$ is computed as:
\begin{equation}
  \mathbb{E}[V_{n-1}(Y)] = \int_0^\infty V_{n-1}(y) \, \theta e^{-\theta y} \, dy  
\end{equation}

The value function \(V_{n-1}(y)\) can be divided into two parts, based on the stopping threshold \(\tau^*\):
\begin{equation}
V_{n-1}(y) = 
\begin{cases}
y, & \text{if} \quad y \leq \tau^* \\
\mathbb{E}[V_{n-2}(Y)], & \text{if} \quad y > \tau^*
\end{cases}
\end{equation}

Therefore, the expectation is split into two parts:
\begin{equation}
\mathbb{E}[V_{n-1}(Y)] = \int_0^{\tau^*} y \, \theta e^{-\theta y} \, dy + \int_{\tau^*}^\infty \mathbb{E}[V_{n-2}(Y)] \, \theta e^{-\theta y} \, dy
\end{equation}

Using integration by parts, let \(u = y\) and \(dv = \theta e^{-\theta y} \, dy\). Then:
\begin{equation}
du = dy, \quad v = -e^{-\theta y}
\end{equation}
Apply the integration by parts formula \(\int u \, dv = uv - \int v \, du\):
\begin{equation}
\int_0^{\tau} y \theta e^{-\theta y} dy = \left[-y e^{-\theta y} \right]_0^{\tau} + \int_0^{\tau^*} e^{-\theta y} dy
\end{equation}

\begin{equation}
   [-y e^{-\theta y}]_0^{\tau} = -\tau e^{-\theta \tau^*} + 0
\end{equation}

\begin{equation}
    \int_0^{\tau} e^{-\theta y} dy = \left[ -\frac{1}{\theta} e^{-\theta y} \right]_0^{\tau} = \frac{1}{\theta} (1 - e^{-\theta \tau^*})
\end{equation}

Since \(\mathbb{E}[V_{n-2}(Y)]\) is a constant, we factor it out:
\begin{equation}
\int_{\tau^*}^\infty \mathbb{E}[V_{n-2}(Y)] \, \theta e^{-\theta y} \, dy = \mathbb{E}[V_{n-2}(Y)] \int_{\tau^*}^\infty \theta e^{-\theta y} \, dy
\end{equation}
The integral of the exponential PDF from \(\tau^*\) to \(\infty\) is:
\begin{equation}
\int_{\tau^*}^\infty \theta e^{-\theta y} \, dy = e^{-\theta \tau^*}
\end{equation}
So, the second integral becomes:
\begin{equation}
\int_{\tau^*}^\infty \mathbb{E}[V_{n-2}(Y)] \, \theta e^{-\theta y} \, dy = \mathbb{E}[V_{n-2}(Y)] \, e^{-\theta \tau}
\end{equation}

The total expected value \(\mathbb{E}[V_{n-1}(Y)]\) is:
\begin{equation}
\mathbb{E}[V_{n-1}(Y)] = -\tau^* e^{-\theta \tau} + \frac{1}{\theta} (1 - e^{-\theta \tau}) + \mathbb{E}[V_{n-2}(Y)] \, e^{-\theta \tau^*}
\end{equation}

The optimal stopping threshold is found by solving the following equation:
\begin{equation}
W(\tau^*) = \mathbb{E}[V_{n-1}(Y)]
\end{equation}
Since \(W(\tau^*) = \tau^*\), the stopping condition becomes:
\begin{equation}
\tau^* = -\tau^* e^{-\theta \tau} + \frac{1}{\theta} (1 - e^{-\theta \tau}) + \mathbb{E}[V_{n-2}(Y)] \, e^{-\theta \tau^*}
\end{equation}
 As $n \to \infty$, we can assume $\mathbb{E}[V_{n-2}(Y)] \approx \tau^*$, which gives:
\begin{equation}
    \tau^* = \frac{1 - e^{-\theta \tau^*}}{\theta}
\end{equation}

\end{proof}
If we assume that the SADs $Y_1, Y_2, \ldots, Y_N$ are i.i.d. samples from an $\text{Exponential}(\theta)$ distribution, the stopping rule is simplified to:
\[
\tau^* = \min \left\{ k \, \middle| \, Y_k \leq -\frac{\log(\delta)}{\theta} \right\}
\]
We can summarize the aforemntioned in the following: as soon as a SAD value is smaller than the threshold -$\frac{\log(\delta)}{\theta}$, we stop. An instance of algorithm from the model of Adaptive ME shown in Figure \ref{fig:AME} is presented below. The complexity of the Adpative ME in the worst case scenario where the loop goes up to N, i.e. there is no stopping rule, is the $O(N^2)$.

\begin{algorithm}
\caption{Adaptive ME for SAD-Based Search}
\label{alg:adaptive_me_sad}
\begin{algorithmic}[1]
\State Initialize $\mathcal{S} \gets \emptyset$
\State Set $\delta \gets 0.05$ \Comment{Stopping threshold}
\State Set $stopping\_time \gets 0$

\For{$k = 1$ to $N$}
    \State $SAD_k \gets \mathrm{calculate\_SAD}(current\_block, candidate\_block[k])$
    \State $\mathcal{S} \gets \mathcal{S} \cup \{SAD_k\}$
    
    \State $F_Y(SAD_k) \gets 
    \dfrac{\left|\{s \in \mathcal{S}: s \leq SAD_k\}\right|}{|\mathcal{S}|}$
    
    \If{$F_Y(SAD_k) \geq 1-\delta$}
        \State $stopping\_time \gets k$
        \State \textbf{break}
    \EndIf
\EndFor
\end{algorithmic}
\end{algorithm}

\section{FAST-ME: Foundation-aware Adaptive Stopping for Targeted Motion Estimation} \label{sec:hybrid}

While traditional block motion estimation evaluates candidate matches based on pixel similarity (e.g., via SAD), it lacks awareness of semantic relevance. To extract high-level semantic understanding from video frames, we consider using foundation models trained on large-scale visual corpora:
\begin{itemize}
    \item Vision Transformer (ViT) for attention-aware spatial representation.
    \item Segment Anything Model (SAM) to extract object masks that highlight salient regions.
    \item CLIP for aligning visual features with natural language prompts or action labels.

\end{itemize}
All models are used to extract attention maps and derive a semantic importance score $ A_k \in [0,1]$ for each candidate motion block. The proposed framework is not algorithm-constrained. 

\subsection{ Semantic Feature Extraction}
\begin{figure}
    \centering
    \includegraphics[width=0.75\linewidth]{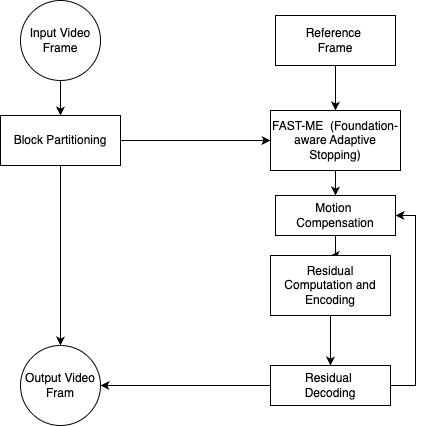}
    \caption{Fast-ME Model diagram }
    \label{fig:fastme}
\end{figure}

The FAST-ME (Foundation-aware Adaptive Stopping for Targeted Motion Estimation) is a hybrid motion estimation framework that integrates semantic cues from foundation models with classical block-matching and Optimal Stopping Theory (OST). The FAST-ME algorithm as shown in Figure \ref{fig:fastme} operates by sequentially evaluating candidate motion vectors within a block’s search window while dynamically integrating both low-level pixel similarity and high-level semantic cues. For each candidate, it computes a blended cost function that down-weights the importance of low-salience regions and imposes tighter matching requirements on semantically meaningful areas. The stopping condition is adjusted dynamically based on attention scores derived from foundation models. As a result, blocks in the background or with low semantic content are likely to terminate early, conserving computational resources. In contrast, objects of interest—such as humans, vehicles, or high-motion foreground elements—trigger more rigorous searches, ensuring their motion is estimated with greater accuracy. This asymmetric search strategy enables FAST-ME to reduce unnecessary block comparisons while maintaining visual quality, making it especially suitable for real-time or resource-constrained video applications.

Each video frame is processed through the foundation model to obtain a semantic relevance score for each block. This score reflects how “important” or “salient” a block is from a scene understanding perspective. The attention weight $A_k$ for candidate block $k$ is then combined with the original distortion measure (SAD value $Y_k$) to create a blended cost:
\begin{equation}
  Y_k = \alpha \cdot Y_k + (1 - \alpha) \cdot (1 - A_k)
  \end{equation}

Here, $\alpha \in [0,1]$ is a tunable hyperparameter controlling the balance between visual fidelity and semantic importance. A lower $Y_k$ suggests a better match that is also semantically meaningful, guiding the adaptive search more intelligently.

\begin{figure}
    \centering
    \includegraphics[width=0.75\linewidth]{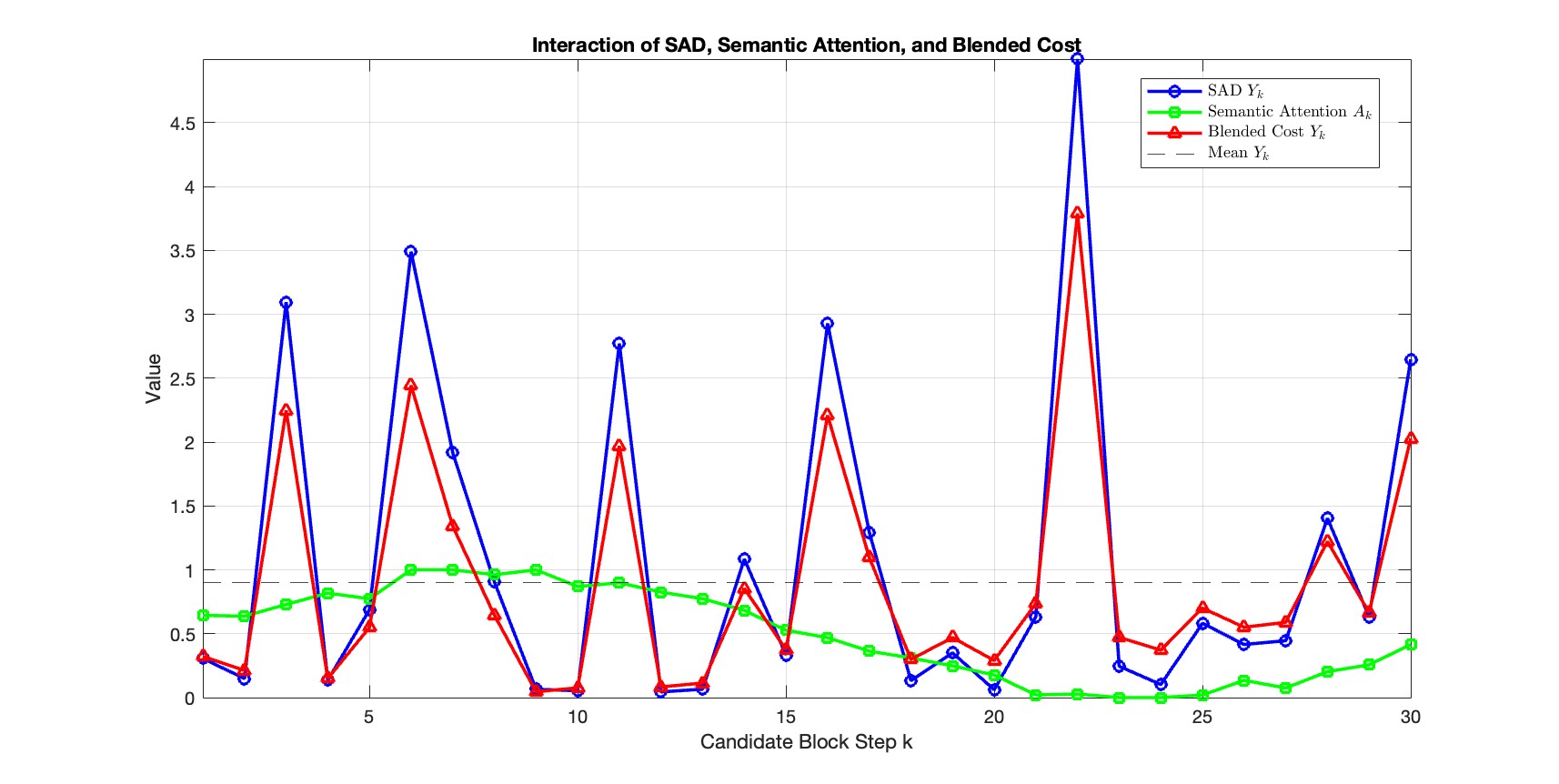}
    \caption{Interaction of SAD, Semantic Attention, and Blended Cost}
    \label{fig:sad_semantic_attection}
\end{figure}

Figure \ref{fig:sad_semantic_attection} illustrates this interaction between the SAD values, semantic attention scores, and the resulting hybrid cost across candidate blocks.

\subsection{Problem Reformulation with Semantic Attention}
The FAST-ME algorithm operates on a per-block basis across each frame in a video sequence. For each block \( B_k \) in the current frame \( I_t \), the algorithm iteratively evaluates candidate displacements within a predefined search window \( [-p, p]^2 \). At every candidate location, the Sum of Absolute Differences (SAD) value \( Y_k \) is computed between the target block and the corresponding block in the reference frame \( I_{t-1} \). Simultaneously, a semantic attention score \( A_k \in [0,1] \) is retrieved from a precomputed attention map generated by a foundation model (e.g., ViT or SAM). These values are fused into a blended cost \( \tilde{Y}_k \) using a convex combination controlled by a tunable parameter \( \alpha \in [0,1] \), as follows:
\[
\tilde{Y}_k = \alpha \cdot Y_k + (1 - \alpha) \cdot (1 - A_k)
\]

The attention score also modulates the search termination condition via a dynamic threshold:
\[
\delta_k = \delta_0 (1 - A_k), \quad T_k = -\frac{\log(\delta_k)}{\theta}
\]
where $\delta_0$ is a global baseline probability (e.g., 0.05).
A candidate match is accepted—and the search terminated—when \( \tilde{Y}_k \leq T_k \) and the associated \( Y_k \) is lower than the best SAD observed so far. This mechanism allows the algorithm to terminate early in visually redundant or background regions (low \( A_k \)), while enforcing stricter matching in semantically important regions (high \( A_k \)). As a result, FAST-ME intelligently allocates computational effort, achieving efficient motion estimation that preserves visual and contextual fidelity.

In more details, let $Y_k$ denote the SAD value of the $k$-th candidate block. Let $A_k \in [0, 1]$ be the \textit{semantic attention score} for that block, where a higher $A_k$ indicates greater semantic relevance. We define a new blended cost function:

\begin{equation}
    \tilde{Y}_k = \alpha \cdot Y_k + (1 - \alpha) \cdot (1 - A_k)
\end{equation}

When $A_k$ is high (semantically salient), $(1 - A_k)$ becomes small, allowing $Y_k$ to dominate. In contrast, low-attention regions increase $\tilde{Y}_k$, favoring early termination.

We assume that $\tilde{Y}_k$ follows an exponential distribution with parameter $\theta$, so the CDF is:

\begin{equation}
    F_{\tilde{Y}}(\tilde{Y}_k) = 1 - e^{-\theta \tilde{Y}_k}
\end{equation}

 The optimal stopping rule becomes:

\begin{equation}
    \tau^* = \min \left\{ k \ \middle| \ F_{\tilde{Y}}(\tilde{Y}_k) \geq 1 - \delta_k \right\}
\end{equation}

Solving for $\tilde{Y}_k$ yields the adaptive stopping boundary:

\begin{equation}
    \tilde{Y}_k \geq -\frac{\log(\delta_0 (1 - A_k))}{\theta}
\end{equation}

The following algorithm summarizes the hybrid OST process for motion estimation:

\begin{algorithm}[H]
\caption{FAST-ME: Semantic-Aware Motion Estimation with Optimal Stopping}
\begin{algorithmic}[1]
\Require Current frame $I_t$, reference frame $I_{t-1}$, block size $B$, search range $p$, semantic attention scores $A_k$, parameters $\alpha$, $\delta_0$, $\theta$
\Ensure Motion vector $(dx^*, dy^*)$ for each block
\For{each block $k$ in $I_t$}
    \State Extract block $B_k$ at location $(x,y)$
    \State Initialize $Y_k^{\min} \gets \infty$
    \For{each candidate displacement $(dx, dy)$ within $[-p, p]^2$}
        \State Compute SAD: $Y_k \gets \text{SAD}(B_k, B_{\text{ref}})$
        \State Retrieve $A_k$ from semantic attention map
        \State Compute blended cost: $\tilde{Y}_k \gets \alpha \cdot Y_k + (1 - \alpha) \cdot (1 - A_k)$
        \State Compute adaptive threshold: $\delta_k \gets \delta_0 (1 - A_k)$
        \State Compute stopping boundary: $T_k \gets -\log(\delta_k)/\theta$
        \If{$\tilde{Y}_k \leq T_k$ \textbf{and} $Y_k < Y_k^{\min}$}
            \State Update $Y_k^{\min}$ and best motion vector $(dx^*, dy^*)$
            \State \textbf{break}
        \EndIf
    \EndFor
    \State Assign motion vector $(dx^*, dy^*)$ to block $k$
\EndFor
\end{algorithmic}
\end{algorithm}

\section{Experiments}\label{sec:experiments}
\begin{figure}
    \centering
    \includegraphics[width=0.6\linewidth]{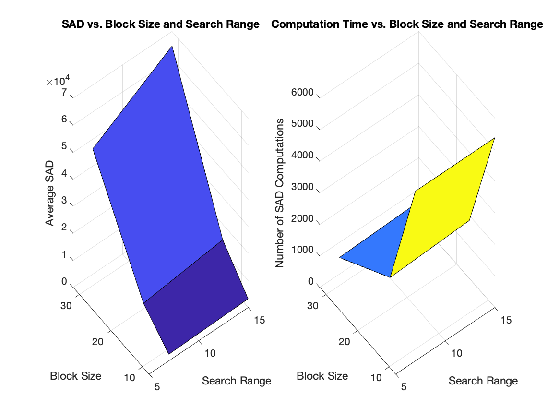}
    \caption{Adaptive ME simulation based on different block and search range}
    \label{fig:SAD_B_S}
\end{figure}

In this section, we report an experimental evaluation to compare the performance of the proposed framework of adaptive motion estimation. To fine-tune the Adaptive ME model, we are investigating the performance of the models based on different block sizes $b$ and the search range $p$. Smaller block sizes ($N=8$, ) have lowest SAD values which is expected as smaller blocks allow finer motion estimation and larger blocks ($N=32$) result in higher SAD values, suggesting that large blocks might not be able to accurately capture motion, especially for fast-moving or small objects. In Figure \ref{fig:SAD_B_S}, it is presented the effect of differences in search range on SAD, where increasing the search range from $±7$ to $±15$ generally increases the SAD, particularly for larger blocks. The outcome shows that a wider search range improves the match accuracy for larger blocks but not as significantly for smaller blocks ($N=8$). To evaluate the performance of the proposed frameworks, we conduct experiments on a diverse set of video datasets:

\begin{itemize}
    \item \textbf{Foreman}, \textbf{BridgeFar} from the Xiph.org DERF dataset — commonly used for motion estimation benchmarking.
    \item \textbf{Four People}, \textbf{Big Buck Bunny} — high-resolution scenes with rich motion and variable semantic content.
\end{itemize}

These datasets \cite{BlockMatching13} and \cite{lai2019} represent a wide variety of motion types, spatial resolutions, and semantic densities, making them suitable for comprehensive evaluation.

The experimental setup has as follows: We use the motion estimation techniques, i.e. Full Search (FS), Diamond Search (DS) and Three-step search (TTS), as provided in \cite{BlockMatching13} implemented in MATLAB. The input media dataset is also used by the mjpegtools project \cite{DerfData}. Semantic attention scores $A_k{model}$ are extracted using a pretrained model $\in{VIT, SAM, CLIP}$ via the HuggingFace \texttt{transformers} library. Each frame is resized to $224 \times 224$, and attention scores are aggregated over $16 \times 16$ blocks to match the motion estimation grid. We have set for FAST-MR the blending factor $\alpha = 0.7$, threshold base: $\delta_0 = 0.05$
 exponential rate: $\theta = 1.0$. We provide a comparison on the blending factor $\alpha$, which controls the trade-off between distortion (SAD) and semantic saliency below. As shown in Table \ref{tab:ablation}, increasing $\alpha$ improves motion fidelity (resulting in lower SAD), while semantic prioritization decreases (lower SCS). This confirms that $\alpha$ enables tunable control over the motion-saliency trade-off. 

\begin{table}[ht]
\centering
\caption{Impact of $\alpha$ on FM Performance}
\begin{tabular}{cccc}
\toprule
$\alpha$ & Mean SAD  & SCS (\%) Comparisons \\
\midrule
0.3 & 11214 & 76.5 & 980 \\
0.5 & 9812  & 72.4 & 1150 \\
0.7 & 9012  & 69.7 & 1320 \\
0.9 & 8856  & 61.0 & 1670 \\
\bottomrule
\end{tabular}
\label{tab:ablation}
\end{table}
 
\begin{figure}
    \centering
    \includegraphics[width=0.7\linewidth]{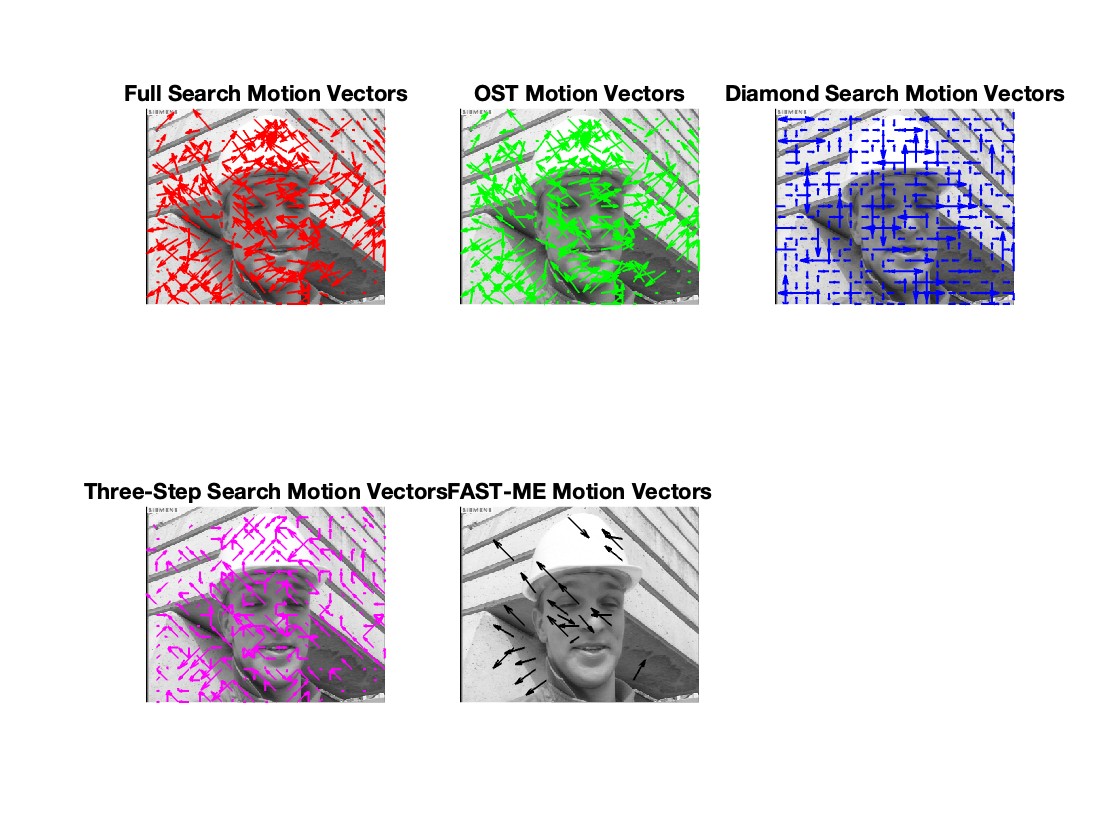}
    \caption{Motion Vector Comparison with different functions}
    \label{fig:mv_comparison}
\end{figure}

In Tables \ref{tab:alg_foreman}, \ref{tab:results} we present the performance comparison of all techniques and various adaptations of OST for various $T$ values for a specific video, i.e. Foreman video. The comparison of the techniques is based on SAD Accuracy, i.e. the average SAD for each approach for multiple executions of the techniques, and number of comparisons needed. The motion vectors are presented in Figure \ref{fig:mv_comparison}. As the threshold $T$ controls how strict the stopping rule is, we provide further insights of Adaptive ME. The FAST-ME method achieves in both tables \ref{tab:alg_foreman}, \ref{tab:results} the lowest SAD among fast approaches, while reducing the number of comparisons by over 98\% compared to Full Search. It also achieves the highest semantic coverage, i.e. percentage of motion vectors that intersect with the top 20\% most semantically important blocks, indicating effective alignment with content relevance. 

\begin{table}
\centering
\begin{tabular}{lcccc}
\toprule
      \textbf{ Algorithm} &\textbf{Time} & \textbf{Avg SAD} ({\it{t}}) & \textbf{N. Comparisons}\\
        \midrule
        FS & 0.0709 &  4.6036e+05 &  80896\\
DS  & 2.1754 & 1.5527e+04 &  4391\\
TSS & 4.8145  &  4.1761e+04 &  10102\\
Adaptive ME T=0 & 0.0420 & 4.6036e+05 & 80896 \\
Adaptive ME T=20 & 0.0032 & 1.0426e+04 & 1631 \\
Adaptive ME T=50 & 0.0130 & 1.0094e+04 & 1495\\
Adaptive ME T=100& 0449 & 9.3964e+03 & 1331\\
\textbf{FAST ME} & 0,0013 &  9.012e+03& 1320 \\
\bottomrule
    \end{tabular}
    \caption{Performance evaluation of ME approachea}
    \label{tab:alg_foreman}
\end{table}

\begin{table}
\centering
\begin{tabular}{lcccc}
\toprule
\textbf{Algorithm} & \textbf{PSNR}   & \textbf{Comparisons} $\downarrow$ & \textbf{SCS (\%)} $\uparrow$ \\
\midrule
Full Search     & 27.89   & 80896 & 45.6 \\
TSS              & 24.82  & 10102 & 50.3 \\
Diamond Search   & 26.89   & 4391  & 52.1 \\
Adaptive ME         & 25.43   & 1495  & 55.2 \\
\textbf{FAST-ME}  & \textbf{26.12} & \textbf{1320} & \textbf{69.7} \\
\bottomrule
\end{tabular}
\caption{Performance on Foreman Sequence}
\label{tab:results}
\end{table}

\begin{figure}
    \centering
    \includegraphics[width=0.75\linewidth]{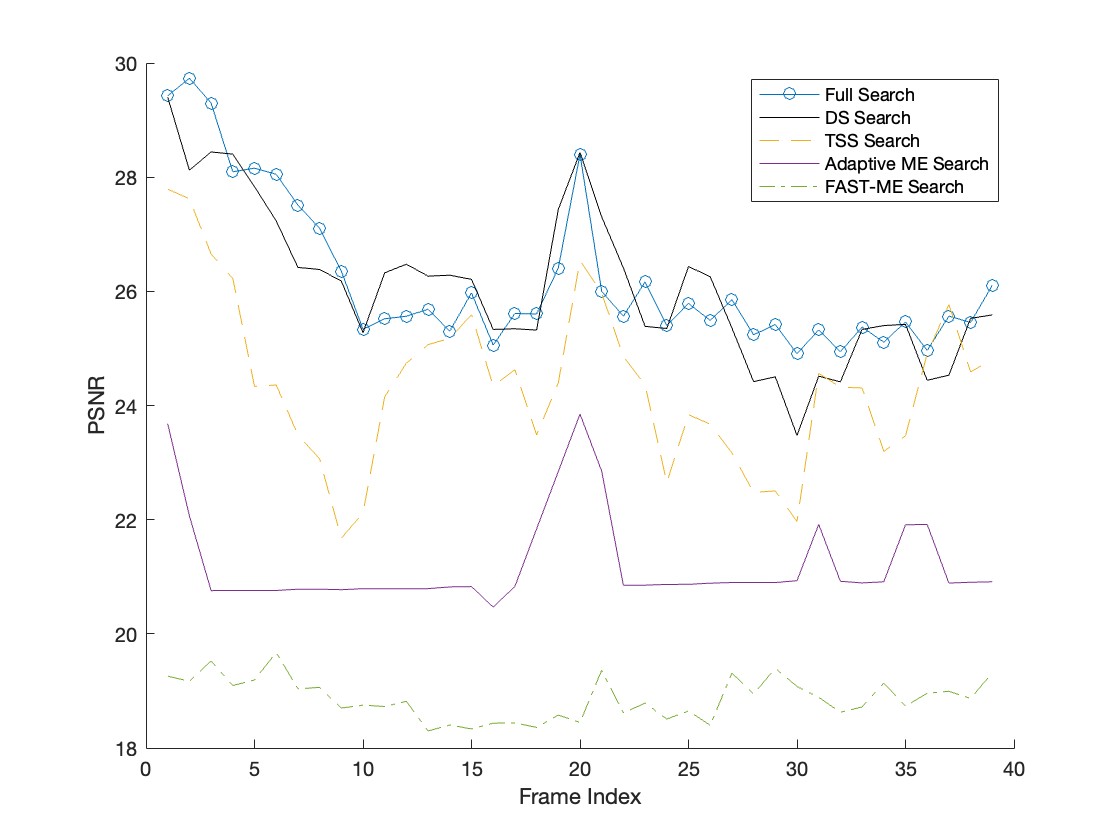}
    \caption{PSNR performance of FS, DS, TSS and Adaptive ME}
    \label{fig:psnr}
\end{figure}

We can overview in Table \ref{tab:alg_foreman} that as $T$ values increase, the SAD values decrease together with the computations number, i.e. the algorithm searches more and finds better matches. However, the improvement in match quality becomes smaller as $T$ value increase. It is worth noting that for $T = 100$, the number of computations is significantly reduced (from $80896$ to $1331$), showing that the search stops early in most cases. The outperformance of Adaptive ME also compared to Peak signal-to-noise ratio (PSNR) can be studied in in Figure \ref{fig:psnr} where the mean time value for FS is $27.89$, for Adaptive ME is  $22.54$, for DS $26.89$ and for TSS is $24.82$.

\begin{table}[h]
    \centering
    \begin{tabular}{cccccc}
    \toprule
        Sequence & Pixel  & ME         & Time  & SAD        & N.Comp \\
    \midrule
         &        & FS         & 1.04  & 469236.80 & 80896 \\
         Foreman & 352x288 & DS         & 3.41  & 16134.95  & 4408 \\
         &        & TSS        & 6.74  & 42593.86  & 10052 \\
         &        & Ad. ME     & 0.0012 & 9939.62   & 1536 \\
         &        & FAST-ME    & 0.68  & 10933.60  & 2300 \\
    \midrule
         &        & FS         & 1.03  & 197880.98 & 58058 \\
         BridgeFar & 352x288 & DS         & 2.08  & 5040.08   & 2871 \\
         &        & TSS        & 3.70  & 14786.33  & 7161 \\
         &        & Ad. ME     & 0.0010 & 4093.91   & 864 \\
         &        & FAST-ME    & 0.42  & 4503.30   & 1296 \\
    \midrule
         &        & FS         & 1.38  & 388611.60 & 783946 \\
         Four People & 1280x720 & DS         & 29.47 & 12443.64  & 38557 \\
         &        & TSS        & 45.51 & 33047.10  & 95156 \\
         &        & Ad. ME     & 0.010 & 10367.34  & 14410 \\
         &        & FAST-ME    & 5.89  & 11404.10  & 21615 \\
    \midrule
         &        & FS         & 1.02  & 129115.68 & 44330 \\
         Big Bunny & 1920x1080 & DS         & 1.36  & 4629.61   & 1974 \\
         &        & TSS        & 3.04  & 13686.54  & 5578 \\
         &        & Ad. ME     & 0.0005 & 1329.02   & 528 \\
         &        & FAST-ME    & 0.27  & 1461.90   & 792 \\
    \bottomrule
    \end{tabular}
    \caption{Performance evaluation between FS, DS, TSS, Adaptive ME, and FAST-ME based on time, SAD, and number of computations across different video sequences}
    \label{tab:all_algo}
\end{table}

\begin{figure}
    \centering
    \includegraphics[width=1.0\linewidth]{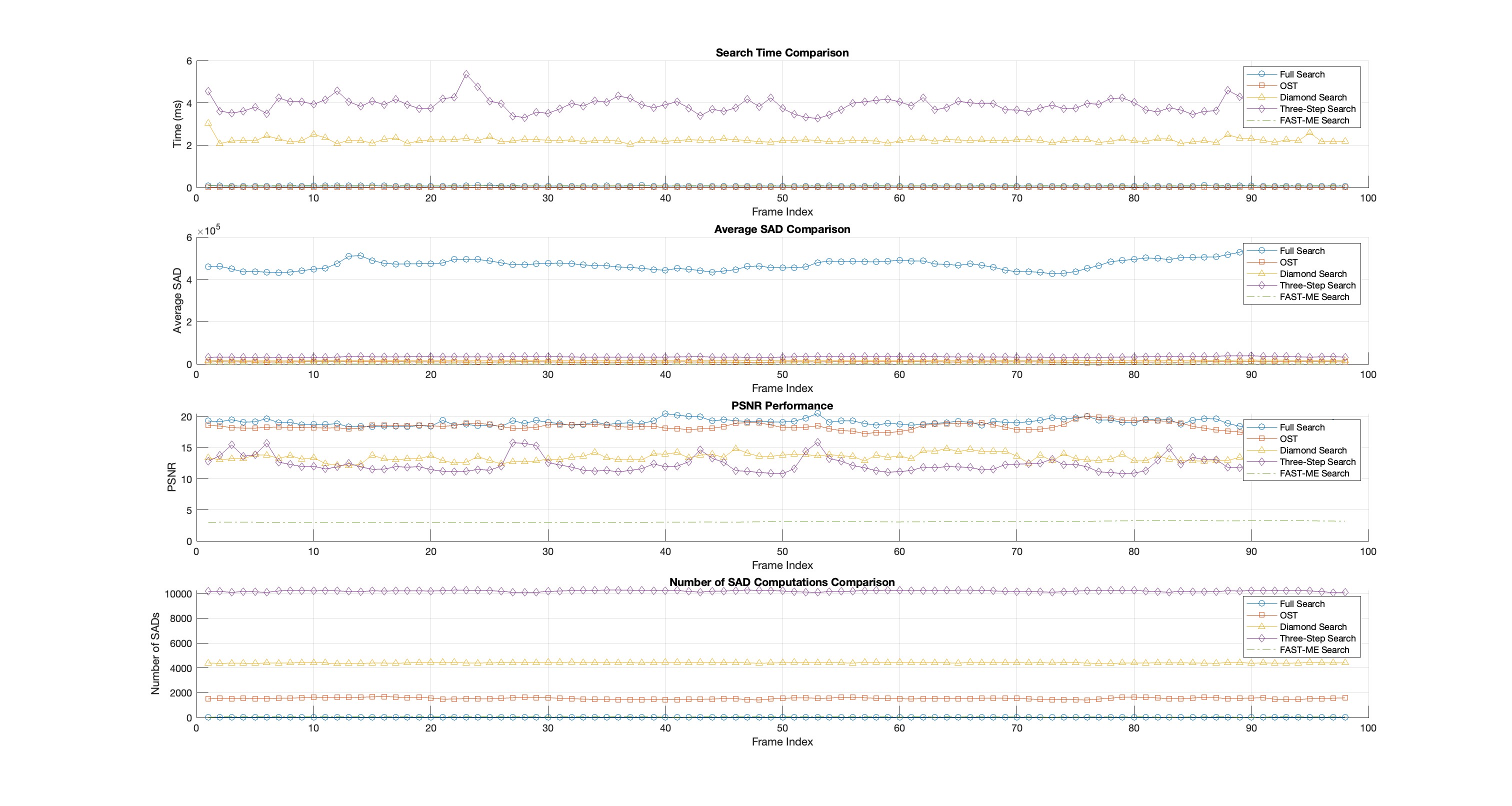}
    \caption{Performance evaluation between FS,DS,TTS, Adaptive ME and Fast-ME based on time, SAD and number of computations}
    \label{fig:fs}
\end{figure}

Going a step further, we conducted $N=100$ runs across frames from multiple video sequences of varying resolutions to evaluate the performance of FS, DS, TSS, Adaptive ME (T=50), and the newly introduced FAST-ME, as shown in Table~\ref{tab:all_algo}. A visual comparison for one representative sequence is illustrated in Figure~\ref{fig:fs}. Full Search (FS) provides the most accurate motion vectors but at a high computational cost. Adaptive ME, on the other hand, offers the best balance between speed and accuracy, reducing the number of computations to just $1$--$2\%$ of FS and execution time by over $99.9\%$, while keeping SAD values within $1$--$3\%$ of FS. Diamond Search (DS) and Three-Step Search (TSS) serve as intermediate approaches. DS achieves lower SAD values than TSS, but is 2--3 times slower than FS due to its larger search pattern. TSS provides moderate accuracy and computation savings but is generally slower than both Adaptive ME and DS.

The newly introduced FAST-ME method demonstrates exceptional efficiency. It achieves the lowest execution time among all tested methods and drastically reduces the number of SAD computations to less than $5\%$ of FS. Notably, for high-resolution videos such as *Four People* and *Big Bunny*, Adaptive ME and FAST-ME offer even greater speed-ups over FS, with substantial reductions in both computation and execution time.

As summarized in Tables~\ref{tab:alg_foreman} and~\ref{tab:all_algo}, DS produces the most accurate motion vectors with the lowest average SAD ($\approx 1.55 \times 10^4$). Adaptive ME follows with a significantly lower SAD than FS ($\approx 1.73 \times 10^5$), offering a practical trade-off between speed and quality. TSS, while less accurate than DS and Adaptive ME, still outperforms FS. FAST-ME delivers the highest speedup, and although its SAD is comparatively higher, its ultra-low computation makes it highly attractive for real-time video processing.

\begin{figure}[ht]
    \centering
    \includegraphics[width=0.8\linewidth]{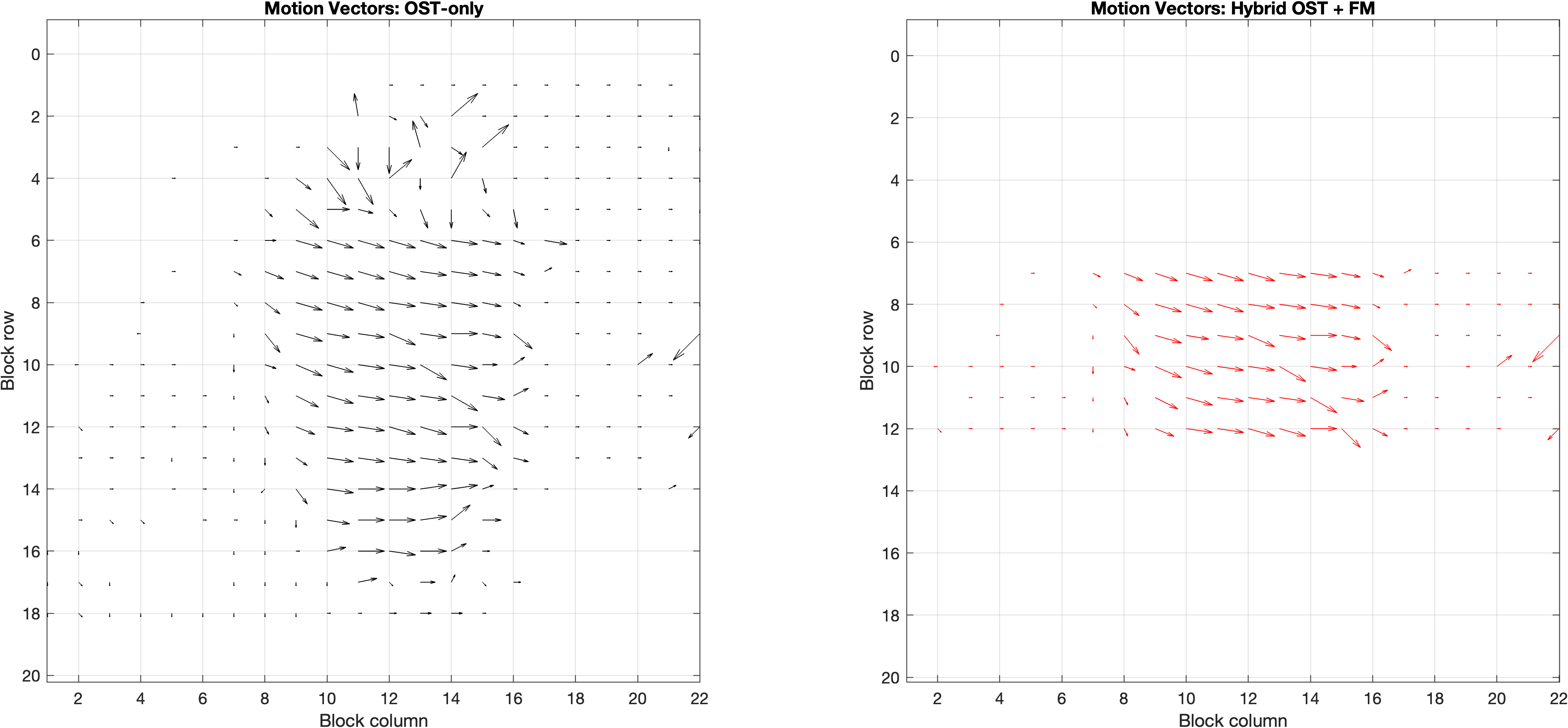}
    \caption{Motion vector fields. Left: OST-only. Right: Hybrid OST+FM. Semantic focus is more pronounced on the right.}
   \label{fig:motion_vectors}
\end{figure}

Figure~\ref{fig:motion_vectors} compares motion vector fields generated by the Adaptive NE and the proposed FAST-ME framework. The latest concentrates vectors in semantically important areas, such as faces and moving limbs, whereas the baseline distributes motion uniformly.

\section{Conclusion and Future Work}\label{sec:conclusion}

In this paper, we introduced \textbf{FAST-ME} (Foundation-aware Adaptive Stopping for Targeted Motion Estimation), a hybrid motion estimation framework that integrates semantic attention from Foundation Models (FMs) into a probabilistically driven motion estimation process based on Optimal Stopping Theory (OST). FAST-ME adaptively adjusts the motion estimation process by leveraging semantic priors from models such as ViT and SAM to focus computational effort on semantically meaningful regions of video frames.

The proposed FAST-ME framework unifies the strengths of statistical decision-making and deep semantic perception. It introduces a blended cost function combining distortion (SAD) and semantic attention, and formulates a dynamic stopping rule that enables early termination of unnecessary block searches. Experimental results on diverse video datasets demonstrate that FAST-ME reduces computation by more than 98\% compared to Full Search, while maintaining or improving reconstruction quality (e.g., PSNR) and semantic alignment (via Semantic Coverage Score).

The integration of semantic awareness into low-level video processing demonstrates a viable approach to bridging traditional motion estimation and modern video understanding. This positions the approach as a computationally efficient and semantically informed motion estimation strategy for applications in wireless multimedia sensor networks (WSMNs), real-time IoT video analytics, and content-aware compression.

Future directions include the additional integration of audio, text, or scene metadata alongside video semantics to guide motion estimation in multimodal scenarios. All the scenarios can be further tested producing a benchmarking of FAST-ME on embedded platforms (e.g., Jetson, Raspberry Pi) to evaluate performance in power-constrained environments. The hybridization of OST and foundation models opens promising research directions at the intersection of efficient computation, semantic understanding, and real-time video processing.

\bibliographystyle{IEEEtran}

\end{document}